\documentclass[runningheads]{llncs}
\usepackage{graphicx}
\usepackage{booktabs}
\usepackage{multirow}
\usepackage[normalem]{ulem}
\usepackage[sectionbib, numbers]{natbib}
\useunder{\uline}{\ul}{}

\usepackage{amsmath}
\usepackage{framed}
\usepackage{mdframed}
\usepackage{lipsum}

\usepackage{hyperref}

\usepackage{footnote}
\usepackage{enumitem}
\usepackage{longtable}
\usepackage{color}
\usepackage[table,dvipsnames]{xcolor}
\usepackage{adjustbox}
\newcommand{\repeatthanks}{\textsuperscript{\thefootnote}}

\usepackage{diagbox}
\usepackage{array}
\newcolumntype{P}[1]{>{\centering\arraybackslash}p{#1}}
\usepackage{makecell}

\usepackage{xcolor}

\usepackage{amsfonts}

\begin{document}

\title{Robust Educational Dialogue Act Classifiers with Low-Resource and Imbalanced Datasets}
\titlerunning{Robust Dialogue Act Classifiers with Low-Resource and Imbalanced Dataset}







%
%

\author{Jionghao Lin\inst{1, 2}, 
Wei Tan\inst{1, }\thanks{Corresponding author.}, 
Ngoc Dang Nguyen\inst{1, }\repeatthanks, 
David Lang\inst{3},
Lan Du\inst{1},
Wray Buntine\inst{1,4},
Richard Beare\inst{1,5},
Guanliang Chen\inst{1},
and Dragan Ga\v{s}evi\'c\inst{1}}

\authorrunning{J. Lin et al.}

\institute{Monash University, Clayton, Australia \\
\email{\{jionghao.lin1, wei.tan2, lan.du, dan.nguyen2, richard.beare, guanliang.chen, dragan.gasevic\}@monash.edu} \and
Carnegie Mellon University, Pittsburgh, USA \and
Stanford University, Stanford, USA \\
\email{dnlang86@stanford.edu}  \and VinUniversity, Hanoi, Vietnam \\
\email{wray.b@vinuni.edu.vn} \\
\and Murdoch Children's Research Institute \\
\email{richard.beare@mcri.edu.au}
\\
}

%
%
\maketitle              
\vspace{-5mm}

\begin{abstract}

Dialogue acts (DAs) can represent conversational actions of tutors or students that take place during tutoring dialogues. Automating the identification of DAs in tutoring dialogues is significant to the design of dialogue-based intelligent tutoring systems. Many prior studies employ machine learning models to classify DAs in tutoring dialogues and invest much effort to optimize the classification accuracy by using limited amounts of training data (\textit{i.e.,} low-resource data scenario). However, beyond the classification accuracy, the robustness of the classifier is also important, which can reflect the capability of the classifier on learning the patterns from different class distributions. We note that many prior studies on classifying educational DAs employ cross entropy (CE) loss to optimize DA classifiers on low-resource data with imbalanced DA distribution. The DA classifiers in these studies tend to prioritize accuracy on the majority class at the expense of the minority class which might not be robust to the data with imbalanced ratios of different DA classes. To optimize the robustness of classifiers on imbalanced class distributions, we propose to optimize the performance of the DA classifier by maximizing the area under the ROC curve (AUC) score (\textit{i.e.,} AUC maximization). Through extensive experiments, our study provides evidence that (i) by maximizing AUC in the training process, the DA classifier achieves significant performance improvement compared to the CE approach under low-resource data, and (ii) AUC maximization approaches can improve the robustness of the DA classifier under different class imbalance ratios.

\end{abstract}
%
%
%


\section{Introduction}
\setcounter{footnote}{0}

One-on-one human tutoring has been widely acknowledged as an effective way to support student learning \cite{vanlehn2007tutorial, rus2017dialogue, Vail2014IdentifyingEM}. Discovering effective tutoring strategies in tutoring dialogues has been considered a significant research task in the design of dialogue-based intelligent tutoring systems \citep{du2016modelling} (\textit{e.g.,} AutoTutor \citep{nye2014autotutor}) and the practice of effective tutoring \cite{lin2022good, Vail2014IdentifyingEM}. A common approach of understanding the tutoring dialogue is to use dialogue acts (DAs), which can represent the intent behind utterances \cite{Vail2014IdentifyingEM, lin2022good, rus2017analysis}. For example, a tutor's utterance (\textit{e.g.,} ``\textit{Well done!}'') can be characterized as the \texttt{Positive Feedback (FP)} DA \cite{Vail2014IdentifyingEM}. To automate the identification of DAs in tutoring dialogues, prior research has often labeled a limited number of utterances and used these labeled utterances to train machine learning models to classify the DAs \citep{rus2017dialogue, rus2017analysis, nye2015automated, lin2022good}. 





Prior research on educational DA classification has demonstrated promising classification accuracy in reliably identifying DAs \cite{samei2015hierarchical, d2010mining, ezen2015understanding, min2016predicting, lin2022good}. However, the existing studies might overlook the impact of imbalanced DA class distribution in the classifier training process. Many of them (\textit{e.g.,} \cite{d2010mining, ezen2015understanding, min2016predicting, lin2022good}) have limited labeled DA datasets (\textit{i.e.,} low-resource scenario \cite{nguyen2022auc}) and certain types of DA may be the minority DA class in the dataset (\textit{i.e.,} imbalanced scenario \cite{nguyen2022auc}). For example, the DA about feedback is rarely seen in the labeled dataset in the work by Min \textit{et al.} \cite{min2016predicting} but the provision of feedback is an important instructional strategy to support students. We argue that the issue of the imbalanced DA classes on the low-resource dataset might negatively impact the DA classification, especially for identifying crucial but underrepresented classes. To obtain more reliable classified DAs, it is necessary to investigate the approach to enhance the classifier robustness, which involves the capability of the classifier on learning the patterns from different class distributions under the low resources \cite{nguyen2022auc, taori2020measuring}.

A robust classifier is able to maintain the performance when the distribution of classes in input data is varied \cite{taori2020measuring}. The prior works in educational DA classification have optimized the DA classifiers by the \textit{Cross Entropy} (CE) loss function which often tends to prioritize accuracy on the majority class at the expense of the minority class(es) and might not be robust enough to highly imbalanced class distribution \cite{nguyen2022auc}.  Inspired by the recent advances in the model robustness literature, we propose to optimize the \textit{Area Under ROC Curve} (AUC) of the DA classifier. AUC score is a metric that can measure the capability of the classifier in distinguishing different classes \cite{nguyen2022auc, taori2020measuring, yuan2022compositional, auc_survey}. Maximizing the AUC score in the model training process has been shown to benefit the performance of classifiers with low-resource and highly-imbalanced scenarios in many domains (\textit{e.g.,} medical and biological domains) \cite{nguyen2022auc, taori2020measuring}. However, the use of AUC maximization is still under-explored in the educational domain. With the intent to enhance the robustness of the DA classifier, we conducted a study to explore the potential value of the AUC maximization approaches on \textit{(i) the classification performance of a DA classifier under the low-resource scenario} and \textit{(ii) the robustness of the DA classifier on the highly-imbalanced data distribution}.

To demonstrate the effectiveness of AUC maximization, we adapted approaches of AUC maximization to replace the CE loss function for the DA classifier. We compared the classification performance of the DA classifier between the classifier optimized by CE and by AUC maximization on the low-resource and imbalanced DA dataset. Through extensive experiments, our studies demonstrated that (i) the adoption of the AUC Maximization approaches outperformed the CE on the low-resource educational DA classification, and (ii)  the AUC maximization approaches were less sensitive to the impacts of imbalanced data distribution than the CE.

\section{Background}
\subsection{Educational Dialogue Act Classification}
\label{rw_part_1}
Existing research in the educational domain has typically trained machine learning models on the labeled sentences from tutoring dialogues to automate the DA classification \cite{samei2015hierarchical, d2010mining, ezen2015understanding, min2016predicting, lin2022good, samei2014context, boyer2010dialogue}. Boyer \textit{et al.} \cite{boyer2010dialogue} trained a Logistic Regression model on 4,806 labeled sentences from 48 tutoring sessions. Their work \cite{boyer2010dialogue} achieved the accuracy of 63\% in classifying 13 DAs. Samei \textit{et al.} \cite{samei2014context} trained Decision Trees and Naive Bayes models on 210 sentences randomly selected from their full tutoring dialogue corpus and they achieved the accuracy of 56\% in classifying 7 DAs. In the later study, Samei \textit{et al.} \cite{samei2015hierarchical} trained a Logistic Regression model on labeled sentences from 1,438 tutorial dialogues and achieved an average accuracy of 65\% on classifying 15 DAs. Though achieving satisfied performance on DA classification, we argue that these studies overlooked the imbalanced DA class distribution, which could negatively impact the DA classifier performance on minority but crucial classes. For example, in the dataset in \cite{samei2015hierarchical}, 23.9\% sentences were labeled as \texttt{Expressive} (\textit{e.g.,} ``\textit{Got it!}'') while 0.3\% labeled as \texttt{Hint} (\textit{e.g.,} ``\textit{Use triangle rules}''). They obtained Cohen's $\kappa$ score of 0.74 on identifying \texttt{Expressive} but 0.34 on \texttt{Hint}. It should be noted that the provision of a hint is an important instructional strategy in the tutoring process, but Cohen's $\kappa$ score on identifying the hint DA was not sufficient. Therefore, it is necessary to enhance the robustness of the DA classifier on imbalanced data.

\subsection{AUC Maximization on Imbalanced Data Distribution}
\label{rw_part3}

The Area Under ROC Curve (AUC) is a widely used metric to evaluate the classification performance of machine learning classifiers on imbalanced datasets \cite{nguyen2022auc, taori2020measuring, yuan2022compositional, auc_survey}. As discussed, many prior studies in educational DA classification have encountered the challenge of imbalanced DA class distribution \cite{boyer2010dialogue, samei2015hierarchical, min2016predicting}. To enhance the capability of the classifier to the imbalanced data distribution, machine learning researchers have begun employing the AUC maximization approaches to optimize classifiers towards the AUC score \cite{nguyen2022auc, taori2020measuring, yuan2022compositional, auc_survey}. Specifically, the process of AUC maximization aims to optimize the AUC score via an AUC surrogate loss, instead of using the standard objective/loss function (\textit{e.g.,} CE) which aims to optimize the classification accuracy of the classifier. The CE function can support the classifier in achieving sufficient classification performance when labeled training instances are sufficient. However, CE often tends to prioritize accuracy on the majority class at the expense of the minority class \cite{nguyen2022auc}. Thus, the CE loss function is vulnerable to highly imbalanced data distribution in the classifier training process \cite{nguyen2022auc}. It should be noted that most prior research in educational DA classification (\textit{e.g.,} \cite{boyer2010dialogue, samei2015hierarchical, lin2022good}) has used CE to optimize their classifiers on the imbalanced data distribution. As the classifier optimized by AUC maximization approaches might be less sensitive to imbalanced distribution than the CE, we propose that it is significant to explore the potential values of AUC maximization for educational DA classification.

\vspace{-1mm}

\section{Methods}
\vspace{-2mm}
\subsection{Dataset}
\label{sec:methods:subsec:dataset}

In this study, we obtained ethical approval from the Monash Human Research Ethics Committee under ethics application number 26156. The tutorial dialogue dataset used in the current study was provided by an educational technology company that offers online tutoring services. The dataset was collected with the consent of tutors and students for use in research. The dataset included records of tutoring sessions where tutors and students worked together to solve problems in various subjects (\textit{e.g.,} math, physics, and chemistry) through chat-based communication. Our study adopted 50 tutorial dialogue sessions, which contained 3,626 utterances (2,156 tutor utterances and 1,470 student utterances. The average number of utterances per tutorial session was 72.52 (\textit{min} = 11, \textit{max} = 325) where tutors made an average of 43.12 utterances (\textit{min} = 5, \textit{max} = 183) per session and students made an average of 29.40 utterances (\textit{min} = 4, \textit{max} = 142) per session. We provided a sample dialogue in the digital appendix via \url{https://github.com/jionghaolin/Robust}.

\vspace{-2mm}
\subsection{Scheme for Educational Dialogue Act}
To identify the dialogue acts (DAs), our study employed a pre-defined educational DA coding scheme introduced in \cite{Vail2014IdentifyingEM}. This scheme has been shown to be effective for analyzing online one-on-one tutoring dialogues in many studies \cite{ezen2015classifying, lin2022good, vail2016predicting}. The DA scheme \cite{Vail2014IdentifyingEM} was originally designed in a two-level structure. The second-level DA scheme, which included 31 DAs, could present more detailed information from the tutor-student dialogue. Thus, our study decided to use the second-level DA scheme to label the tutor-student utterances. Due to space reasons, we displayed the details of the full DA scheme in an electronic appendix at \url{https://github.com/jionghaolin/Robust}, and the scheme could also be found in \cite{Vail2014IdentifyingEM}. Before the labeling, we first divided each  utterance into multiple sentences as suggested by Vail and Boyer \cite{Vail2014IdentifyingEM} and then we removed the sentences which only contained symbols or emojis. Two human coders were recruited to label the DA for each sentence, and a third educational expert was involved in resolving any disagreements. Two coders achieved Cohen's $\kappa$ score of 0.77, which indicated a substantial level of agreement between the coders.

\subsection{Approaches for Model Optimization}
We aimed to examine the potential values of AUC maximization approaches to enhance the performance of educational DA classifiers. Inspired by the recent works in AUC Maximization \cite{ying2016stochastic,yuan2020large,yuan2022compositional, nguyen2022auc}, 
our study selected the approaches for training the DA classifier as follows:
\begin{itemize}
    \item \textbf{Cross Entropy (CE)}: CE is a commonly used loss function to optimize the classifier. The loss values of CE were calculated by comparing the predicted probability distribution to the true probability distribution. Our study used the DA classifier optimized by the CE approach as the baseline.


    \item \textbf{Deep AUC maximization (DAM)}: Yuan \textit{et al.} \cite{yuan2020large} proposed a robust AUC approach (\textit{i.e.,} DAM) to optimize the performance of the DA classifier by optimizing it for the AUC surrogate loss. Optimizing the AUC score can significantly improve the performance of classifiers on imbalanced data. 
    
    \item \textbf{Compositional AUC (COMAUC)}: COMAUC proposed by Yuan \textit{et al.} \cite{yuan2022compositional} involves minimizing a compositional objective function by alternating between maximizing the AUC score and minimizing the cross entropy loss. Yuan \textit{et al.} \cite{yuan2022compositional} theoretically proved that COMAUC could substantially improve the performance of the classifier. Consequently, this is one of the methods we sought to implement to improve the performance of the DA classifier.
\end{itemize}

\vspace{-5mm}
\subsection{Model Architecture by AUC Maximization}

\begin{figure}[ht]
\centering
  \includegraphics[width=0.8\textwidth]{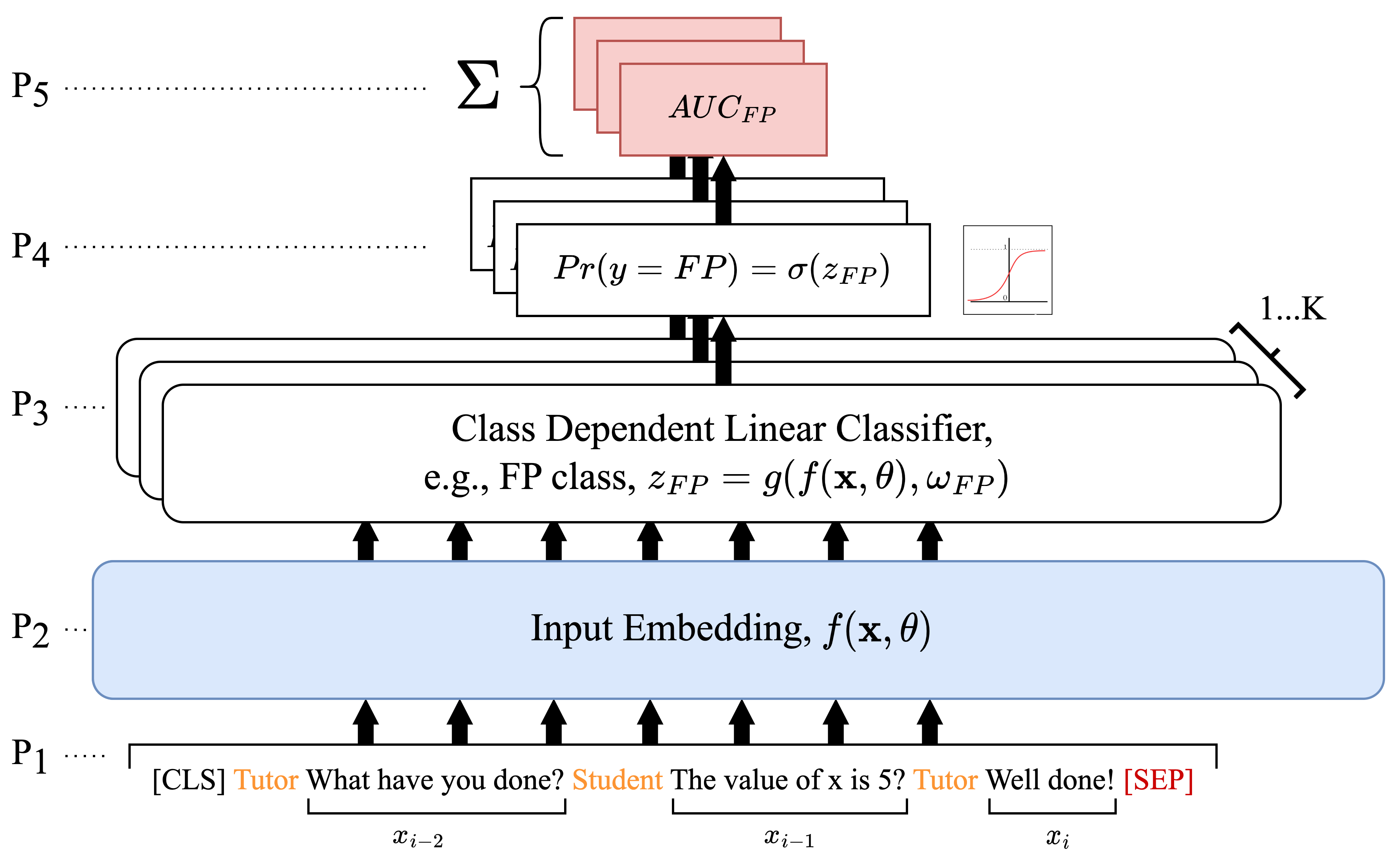}
\caption{Architecture of optimizing the classifier by AUC maximization approaches.}
\label{fig:model_arc}
\vspace{-9mm}
\end{figure}

We denote the training set as $D=\{(x_1, y_1), (x_2, y_2), \ldots, (x_n, y_n)\}$ where $x_i$ represents the $i$-th input utterance with its corresponding label $y_i$, \textit{i.e.,} $y_i \in \{\text{NF}, \text{FP}, \ldots, \text{ACK}\}$ dialogue acts from \cite{Vail2014IdentifyingEM}. $n$ represents the size of our training set, which we assume to be small due to the challenging nature of obtaining data for the DA classification \cite{nye2015automated}. Additionally, we denote $\theta \in \mathbb{R}^{d}$ as the parameters of the utterance encoder $f$, and $\omega  \in \mathbb{R}^{d\times1}$ as the parameters of the class-dependent linear classifier $g$. In \textbf{Phase 1} (P\textsubscript{1} see Fig.~\ref{fig:model_arc}), we input the $i$-th sentence, \textit{i.e.,} $x_i$, along with its contextual sentences \cite{tlt_lin_2022} $x_{i-1}$ and $x_{i-2}$, \textit{i.e.,} $\mathbf{x} = \{{x_i,x_{i-1}, x_{i-2}}\}$, to the utterance encoder. In \textbf{Phase 2} (P\textsubscript{2}), we employed BERT \cite{devlin-etal-2019-bert} as the encoder due to its effectiveness in educational classification tasks \cite{rakovic2022towards, lin2022good, 9970298}. The BERT model can learn the latent encoding representation of the input sentences and output these representations for the class-dependent linear classifier in \textbf{Phase 3} (P\textsubscript{3}). In the following, we illustrate the process of generating the FP\footnote{\textbf{FP}, \texttt{Positive Feedback} ``\textit{Well done!}'', same abbreviation from \cite{Vail2014IdentifyingEM}} prediction for the dialogue $x_i$:

\vspace{-5mm}

{\begin{eqnarray}
z_{\text{FP}}=g\left(\omega_{\text{FP}}, f\left(\theta,\mathbf{x}\right) \right),
\label{eq:latentencoding} \\
\operatorname{Pr}\left( y = \text{FP} \right) = 
\sigma\left(z_{\text{FP}}\right),
\label{eq:probability}
\end{eqnarray}}

Since there are $K$ classes\footnote{Our study has 31 dialogue acts as the classes to be classified}, we need $K$ linear classifiers, each to generate the prediction probability for its corresponding class via using the sigmoid function $\sigma()$ in \textbf{Phase 4} (P\textsubscript{4}). Lastly, to train the model with the surrogate AUC loss, we implement the deep AUC margin loss \cite{yuan2020large} for each of the sigmoid results from \textbf{Phase 4}, we demonstrate this surrogate loss for the \textbf{FP} class as follows:

\begin{equation}
        {\text{AUC}(\text{FP})}=
        \mathbb{E}\left[\left(m
        -\sigma({z_{\text{FP}}})+\sigma\left(z_{\text{FP}}^{\prime}\right)\right)^{2}\right] ,
    \label{eq:AUCLoss}
    \end{equation}

where $\sigma(z_{\text{FP}})$ represents the prediction probability for the sentence that is labeled as \text{FP}, and $\sigma(z_{\text{FP}}^{\prime})$ represents the prediction probability for the sentence that is not. The margin $m$, normally set as $1$ \cite{ying2016stochastic,yuan2020large,yuan2022compositional,nguyen2022auc}, serves to separate the correct and incorrect prediction for the FP class, encouraging the correct and incorrect prediction to be distinguishable from each other. Then, the AUC losses of $K$ classes are collected, summed up and back propagated in \textbf{Phase 5} (P\textsubscript{5}) to tune the BERT encoder parameters $\theta$ and the class-dependent linear classifier parameters $\omega$. We repeat the process until no further improvements can be achieved and obtain the optimal DA classifier for the DA classification.

\subsection{Study Setup}
\label{setup}
We aimed to evaluate the effectiveness of AUC maximization methods in two DA classification scenarios: (i) a low-resource scenario and (ii) an imbalanced distribution scenario. The dataset (50 dialogue sessions) was randomly split to \textit{training} (40 sessions) and \textit{testing set} (10 sessions) in the ratio of \textit{80\%:20\%} where the \textit{training set} contained 3,763 instances and \textit{testing set} contained 476 instances.

\textbf{Low-Resource Scenario.} To examine the efficacy of AUC maximization approaches on the classification performance of the DA classifier, our study first investigated the impact of AUC approaches against the traditional CE for the multi-class DA classification to classify 31 DAs under the low-resource setting. Inspired by \cite{nguyen2022auc}, we simulated the experiments under the low-resource setting with a \textit{training set} size of  $\{25, 50, 100, 200, 400, 800\}$ randomly sampled from the full \textit{training set}. Then, we evaluated the classification performance of the  DA classifier optimized by the CE and AUC approaches on the full \textit{testing set}.  For each \textit{training set} size, we trained the DA classifiers optimized by AUC maximization approaches and the CE baseline on 10 random training partitions and analyzed their average performance. This allowed us to investigate the performance of AUC maximization approaches under low-resource conditions.

\textbf{Imbalanced Scenario.} To study the robustness of the DA classifier to the imbalanced data distribution, we simulated two settings of imbalanced data distribution on the training and testing dataset. For each setting, we conducted binary classification on a specific type of DA and investigated the impact of imbalanced data distribution at different levels. First, we simulated the classifier performance for all approaches with imbalanced data distribution at different levels in the \textit{training set}, which is also known as distribution shifting, \textit{i.e.,} the data distribution on the training set does not match the distribution on testing \cite{nguyen2022auc, taori2020measuring}.  Second, we simulated the data distribution shifting on both \textit{training} and \textit{testing sets}. For both setups, we also developed a random generator that creates imbalanced datasets for the DA classification tasks. This generator created \textit{training} and \textit{testing sets} by sampling sentence sets based on the percentage of specific feedback sentences. By evaluating the performance of AUC maximization methods on these different data distributions, we were able to assess its robustness under various imbalanced data conditions.


\textbf{Evaluation Metrics.}
In line with the prior works in educational DA classification \cite{rus2017analysis, nye2015automated, samei2014context, samei2015hierarchical, lin2022good}, our study also used Cohen's $\kappa$ scores to evaluate the performance of the DA classifier. Additionally, instead of using classification accuracy, we used the F1 score as another measure as the F1 score often presents more reliable results on imbalanced distribution due to its nature of implicitly including both precision and recall performance 
\cite{nguyen2022auc}.

\section{Results}

\subsection{AUC Maximization under Low-Resource Scenario}
\label{RQ1_results}

\begin{figure}[ht]
\centering
  \includegraphics[width=0.9\textwidth]{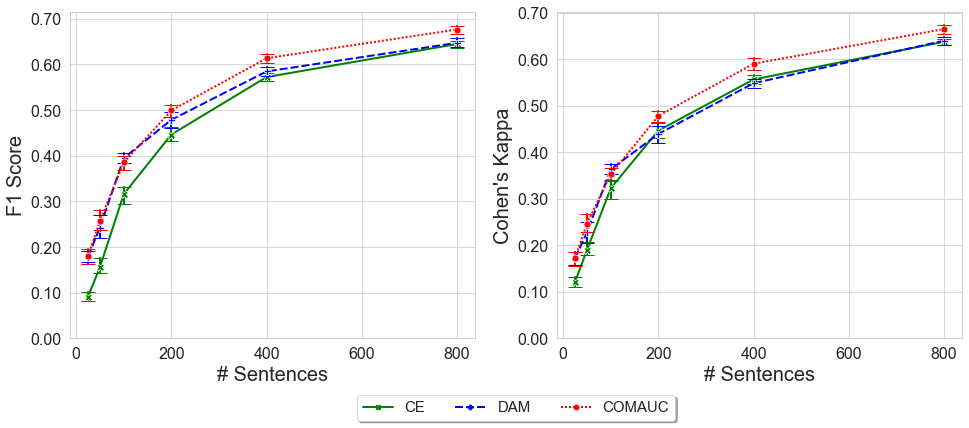}
\caption{Performance of DA classifiers with different optimization approaches.}
\label{fig:RQ1_2}
\vspace{-6mm}
\end{figure}

To evaluate the effectiveness of AUC approaches (\textit{i.e.,} DAM and COMAUC), we investigated the proposed optimization approaches (\textit{i.e.,} CE, DAM, and COMAUC) as described in Sec.\ref{setup}. We run 10 different random seeds for each approach to minimize the impact of random variation and obtain reliable estimations of the model's performance. We plotted the averaged results with error bars for each approach in Fig.\ref{fig:RQ1_2} where the green, blue and red lines represented the approaches CE, AUC, and COMAUC, respectively. Fig. \ref{fig:RQ1_2} shows that when the training set size was small (\textit{e.g.,} 25, 50, 100, 200, 400 sentences), the averaged F1 score of the AUC approaches generally outperformed the CE approach; these differences were significant at the 95\% level of confidence. It should be noted that the gap between AUC and CE approaches achieved the most significant difference at 100 sentences. Furthermore, when the training set size increased to 800 sentences, COMAUC outperformed both DAM and CE on average and demonstrated more stable performance as indicated by the error bars. These findings illustrated that COMAUC is an effective and reliable AUC maximization approach under low-resource scenarios.
\vspace{-3mm}



\subsection{AUC Maximization under Imbalanced Scenario}
To understand the extent to which AUC approaches can enhance the robustness of the DA classifier to the impact of imbalanced data, we conducted binary classification experiments on a DA from two perspectives as discussed in the \textbf{Imbalanced Scenario} in Sec. \ref{setup}. We choose the dialogue act \texttt{Positive Feedback} (FP) as the candidate for analysis as it is widely used in tutoring dialogues.

\begin{figure}[ht]
\centering
  \includegraphics[width=0.9\textwidth]{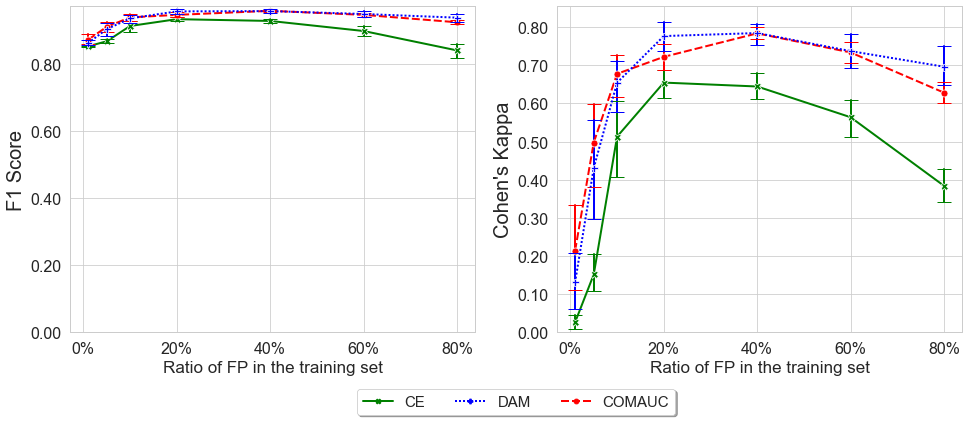}
\caption{Classification performance on FP class with different optimization approaches to the data shift in \textbf{training set}}
\label{fig:RQ21}
\end{figure}

Inspired by the result shown in Fig. \ref{fig:RQ1_2} where the performance gap of F1 score between AUC and CE approaches demonstrated the most significant difference at 100 instances; thus, for the first setting in the imbalanced scenario, we decided to simulate the training set with 100 instances and explored the impact of the distribution shifting on FP class in the training set. As introduced in Sec. \ref{setup}, we adjusted the ratio of FP in the training dataset from 1\% to 80\% and examined the classification performance of the DA classifier in different ratios. In Fig. \ref{fig:RQ21}, we found that the F1 score of the CE approach was lower than those of AUC approaches (\textit{i.e.,} DAM and COMAUC) when the DA classifier was trained on the dataset with the ratio of 80\% for the FP class in the training set. Though the F1 scores of the AUC approaches also decreased under these conditions, the decrease was less pronounced than that of CE. When scrutinizing Cohen's $\kappa$ score, the DA classifier optimized by CE approach demonstrated vulnerable performance when the FP ratios were 60\% and 80\% in the training set whereas the Cohen's $\kappa$ score of both AUC approaches maintained Cohen's $\kappa$ above 0.60.

\begin{figure}[ht]
\centering
  \includegraphics[width=0.9\textwidth]{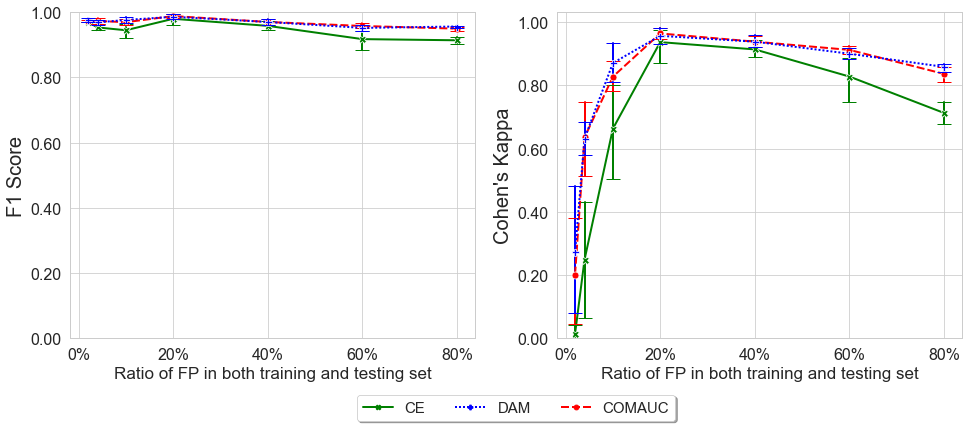}
\caption{Classification performance on FP class with different optimization approaches to the data shift in both \textbf{training} and \textbf{testing set}}
\label{fig:RQ22}
\vspace{-3mm}
\end{figure}

For the second setting in the imbalanced scenario, we also examined the classification performance of the DA classifier on the training set with 100 instances but a smaller testing set that contained 50 instances, which was designed for facilitating the analysis. As introduced in Sec. \ref{setup}, the same ratio of the specific DA class (\textit{e.g.,} FP) was adjusted simultaneously in the training and testing sets. Fig. \ref{fig:RQ22} shows that the three selected approaches achieved promising F1 score in different ratios of FP but the AUC approaches were more stable than the CE approach when the ratios were 60\% and 80\%. When scrutinizing Cohen's $\kappa$ scores, AUC approaches were less susceptible to changes in ratios compared to CE. Additionally, the CE approach demonstrated a higher variance of both the Cohen's $\kappa$ scores and F1 score in Fig. \ref{fig:RQ22} compared to the AUC approaches. These results indicate that the AUC approaches (\textit{i.e.,} DAM and COMAUC) were more robust to the imbalanced data distribution when the ratio of FP was adjusted in the training and the testing set compared to the CE approach.


\section{Discussion and Conclusion}

 Classification is a fundamental task in applying artificial intelligence and data mining techniques to the education domain. However, many educational classification tasks often encounter challenges of low-resource and imbalanced class distribution \cite{al2019predicting, zhao2023mets, cavalcanti2020good, lin2022good} where the low-resource dataset may not be representative of the population \cite{tan2021diversity, tan2023does} and the classifier might overfit the majority class of the dataset \cite{nguyen2022auc}. A robust classifier can optimize the classification performance under low-resourced datasets and provide reliable results from various data distributions \cite{nguyen2022auc}. Thus, enhancing model robustness to low-resourced and imbalanced data is a crucial step to deploying machine learning algorithms in the educational domain. Our study investigated various approaches regarding the robustness of classifiers for the educational DA classification task. Through extensive experiments, our study provided evidence that AUC maximization approaches (\textit{e.g.,} DAM and COMAUC) can enhance the classification performance of DA classifiers on the limited dataset (\textit{i.e.,} low-resource scenario) and the robustness of the classifier on imbalanced data distribution (\textit{i.e.,} imbalanced scenario) compared to the widely-used standard approach, \textit{i.e.,} Cross Entropy.

\noindent\textbf{Implications.} \textit{Firstly}, it is beneficial to adopt the AUC maximization approach for educational DA classification tasks where the training dataset is limited or low-resource. For example, the DA classification in the learning context of medical training \cite{zhao2023mets} also encounters the low-resource issue. Additionally, many classification tasks in the educational domain also encounter the issue of low-resource annotation (\textit{e.g.,} assessment feedback classification \cite{cavalcanti2020good}). Driven by the findings in our study, we suggest adopting the AUC maximization approaches to the low-resource classification tasks in the educational domain. \textit{Secondly}, in real-world tutoring dialogues, the data distribution about DAs from tutors and students is unavoidable to be imbalanced and changeable over time. To obtain a reliable DA classifier, researchers need to fine-tune the DA classifier when the new batch of the training instances is ready. Our results showed that the DA classifier optimized by the widely-used CE approach was brittle to the issues of imbalanced distribution. Additionally, the imbalanced data distribution widely exists in many educational classification tasks. For example, from a practical standpoint, one challenge in predicting student academic performance is the presence of students who are at high risk of failing was highly imbalanced compared to the students who have excellent or medium performance \cite{al2019predicting}. The results of our study call for future research on the potential of AUC approaches in identifying at-risk students from the highly imbalanced data distribution.

\noindent\textbf{Limitations.}
We acknowledged that our study only simulated the imbalanced data distribution by pre-defined imbalanced ratios. It is necessary to investigate the imbalanced data in real-world tutoring dialogues.

\bibliographystyle{splncs04}
\bibliography{mybibliography}

\begin{thebibliography}{10}
\providecommand{\url}[1]{\texttt{#1}}
\providecommand{\urlprefix}{URL }
\providecommand{\doi}[1]{https://doi.org/#1}

\bibitem{al2019predicting}
Al-Luhaybi, M., Yousefi, L., Swift, S., Counsell, S., Tucker, A.: Predicting
  academic performance: {A} bootstrapping approach for learning dynamic
  bayesian networks. In: AIED. pp. 26--36. Springer (2019)

\bibitem{boyer2010dialogue}
Boyer, K., Ha, E.Y., Phillips, R., Wallis, M., Vouk, M., Lester, J.: Dialogue
  act modeling in a complex task-oriented domain. In: Proceedings of the
  SIGDIAL 2010 Conference. pp. 297--305 (2010)

\bibitem{cavalcanti2020good}
Cavalcanti, A.P., Diego, A., Mello, R.F., Mangaroska, K., Nascimento, A.,
  Freitas, F., Ga\v{s}evi\'{c}, D.: How good is my feedback? a content analysis
  of written feedback. In: Proceedings of the LAK. p. 428–437. LAK '20, ACM,
  New York, NY, USA (2020)

\bibitem{devlin-etal-2019-bert}
Devlin, J., Chang, M.W., Lee, K., Toutanova, K.: {BERT}: Pre-training of deep
  bidirectional transformers for language understanding. In: Proceedings of
  NAACL-HLT. pp. 4171--4186. Association for Computational Linguistics,
  Minneapolis, Minnesota (2019)

\bibitem{d2010mining}
D'Mello, S., Olney, A., Person, N.: Mining collaborative patterns in tutorial
  dialogues. Journal of Educational Data Mining  \textbf{2}(1),  1--37 (2010)

\bibitem{du2016modelling}
Du~Boulay, B., Luckin, R.: Modelling human teaching tactics and strategies for
  tutoring systems: 14 years on. International Journal of Artificial
  Intelligence in Education  \textbf{26}(1),  393--404 (2016)

\bibitem{ezen2015understanding}
Ezen-Can, A., Boyer, K.E.: Understanding student language: An unsupervised
  dialogue act classification approach. Journal of Educational Data Mining
  \textbf{7}(1),  51--78 (2015)

\bibitem{ezen2015classifying}
Ezen-Can, A., Grafsgaard, J.F., Lester, J.C., Boyer, K.E.: Classifying student
  dialogue acts with multimodal learning analytics. In: Proceedings of the
  Fifth LAK. pp. 280--289 (2015)

\bibitem{lin2022good}
Lin, J., Singh, S., Sha, L., Tan, W., Lang, D., Ga{\v{s}}evi{\'c}, D., Chen,
  G.: Is it a good move? mining effective tutoring strategies from human--human
  tutorial dialogues. Future Generation Computer Systems  \textbf{127},
  194--207 (2022)

\bibitem{tlt_lin_2022}
Lin, J., Tan, W., Du, L., Buntine, W., Lang, D., Ga{\v{s}}evi{\'c}, D., Chen,
  G.: Enhancing educational dialogue act classification with discourse context
  and sample informativeness. IEEE TLT  (in press)

\bibitem{min2016predicting}
Min, W., Wiggins, J.B., Pezzullo, L.G., Vail, A.K., Boyer, K.E., Mott, B.W.,
  Frankosky, M.H., Wiebe, E.N., Lester, J.C.: Predicting dialogue acts for
  intelligent virtual agents with multimodal student interaction data.
  International Educational Data Mining Society  (2016)

\bibitem{nguyen2022auc}
Nguyen, N.D., Tan, W., Buntine, W., Beare, R., Chen, C., Du, L.: Auc
  maximization for low-resource named entity recognition. In: Proceedings of
  the AAAI Conference on Artificial Intelligence (2023)

\bibitem{nye2014autotutor}
Nye, B.D., Graesser, A.C., Hu, X.: Autotutor and family: A review of 17 years
  of natural language tutoring. International Journal of Artificial
  Intelligence in Education  \textbf{24}(4),  427--469 (2014)

\bibitem{nye2015automated}
Nye, B.D., Morrison, D.M., Samei, B.: Automated session-quality assessment for
  human tutoring based on expert ratings of tutoring success. International
  Educational Data Mining Society  (2015)

\bibitem{rakovic2022towards}
Rakovi{\'c}, M., Sha, L., Nagtzaam, G., Young, N., Stratmann, P.,
  Ga{\v{s}}evi{\'c}, D., Chen, G.: Towards the automated evaluation of legal
  casenote essays. In: International Conference on Artificial Intelligence in
  Education. pp. 167--179. Springer (2022)

\bibitem{rus2017dialogue}
Rus, V., Maharjan, N., Banjade, R.: Dialogue act classification in
  human-to-human tutorial dialogues. In: Innovations in smart learning, pp.
  185--188. Springer (2017)

\bibitem{rus2017analysis}
Rus, V., Maharjan, N., Tamang, L.J., Yudelson, M., Berman, S., Fancsali, S.E.,
  Ritter, S.: An analysis of human tutors’ actions in tutorial dialogues. In:
  The Thirtieth International Flairs Conference (2017)

\bibitem{samei2014context}
Samei, B., Li, H., Keshtkar, F., Rus, V., Graesser, A.C.: Context-based speech
  act classification in intelligent tutoring systems. In: International
  conference on intelligent tutoring systems. pp. 236--241. Springer (2014)

\bibitem{samei2015hierarchical}
Samei, B., Rus, V., Nye, B., Morrison, D.M.: Hierarchical dialogue act
  classification in online tutoring sessions. In: EDM. pp. 600--601 (2015)

\bibitem{9970298}
Sha, L., Raković, M., Lin, J., Guan, Q., Whitelock-Wainwright, A., Gašević,
  D., Chen, G.: Is the latest the greatest? a comparative study of automatic
  approaches for classifying educational forum posts. IEEE Transactions on
  Learning Technologies pp. 1--14 (2022)

\bibitem{tan2021diversity}
Tan, W., Du, L., Buntine, W.: Diversity enhanced active learning with strictly
  proper scoring rules. Advances in Neural Information Processing Systems
  \textbf{34},  10906--10918 (2021)

\bibitem{tan2023does}
Tan, W., Lin, J., Lang, D., Chen, G., Gasevic, D., Du, L., Buntine, W.: Does
  informativeness matter? active learning for educational dialogue act
  classification. In: International Conference on Artificial Intelligence in
  Education. Springer (2023)

\bibitem{taori2020measuring}
Taori, R., Dave, A., Shankar, V., Carlini, N., Recht, B., Schmidt, L.:
  Measuring robustness to natural distribution shifts in image classification.
  Advances in Neural Information Processing Systems  \textbf{33},  18583--18599
  (2020)

\bibitem{vail2016predicting}
Vail, A.K., Grafsgaard, J.F., Boyer, K.E., Wiebe, E.N., Lester, J.C.:
  Predicting learning from student affective response to tutor questions. In:
  ITS. pp. 154--164. Springer (2016)

\bibitem{Vail2014IdentifyingEM}
Vail, A.K., Boyer, K.E.: Identifying effective moves in tutoring: On the
  refinement of dialogue act annotation schemes. In: International conference
  on intelligent tutoring systems. pp. 199--209. Springer (2014)

\bibitem{vanlehn2007tutorial}
VanLehn, K., Graesser, A.C., Jackson, G.T., Jordan, P., Olney, A., Ros{\'e},
  C.P.: When are tutorial dialogues more effective than reading? Cognitive
  science  \textbf{31}(1),  3--62 (2007)

\bibitem{auc_survey}
Yang, T., Ying, Y.: Auc maximization in the era of big data and ai: A survey.
  ACM Comput. Surv.  \textbf{55}(8) (dec 2022). \doi{10.1145/3554729},
  \url{https://doi.org/10.1145/3554729}

\bibitem{ying2016stochastic}
Ying, Y., Wen, L., Lyu, S.: Stochastic {O}nline {AUC} {M}aximization. Advances
  in Neural Information Processing Systems  \textbf{29} (2016)

\bibitem{yuan2020large}
Yuan, Z., Yan, Y., Sonka, M., Yang, T.: Large-scale robust deep {AUC}
  maximization: A new surrogate loss and empirical studies on medical image
  classification. In: 2021 IEEE/CVF ICCV. pp. 3020--3029. IEEE Computer
  Society, Los Alamitos, CA, USA (oct 2021)

\bibitem{yuan2022compositional}
Yuan, Z., Guo, Z., Chawla, N., Yang, T.: Compositional training for end-to-end
  deep {AUC} maximization. In: International Conference on Learning
  Representations (2022), \url{https://openreview.net/forum?id=gPvB4pdu_Z}

\bibitem{zhao2023mets}
Zhao, L., Swiecki, Z., Gasevic, D., Yan, L., Dix, S., Jaggard, H., Wotherspoon,
  R., Osborne, A., Li, X., Alfredo, R., et~al.: {METS}: Multimodal learning
  analytics of embodied teamwork learning. In: LAK23: 13th International
  Learning Analytics and Knowledge Conference. pp. 186--196 (2023)

\end{thebibliography}
%




\end{document}